\documentclass[10pt,twocolumn,letterpaper]{article}

\usepackage{tfm}
\usepackage{times}
\usepackage{epsfig}
\usepackage{graphicx}
\usepackage{amsmath}
\usepackage{bigstrut}
\usepackage{amssymb}
\usepackage{tabularx}
\begin{document}

\title{Comparative study of histogram distance measures for re-identification}

\author{Pedro A. Mar\'in-Reyes, Javier Lorenzo-Navarro, Modesto Castrill\'on-Santana \\
Instituto Universitario SIANI\\
Universidad de Las Palmas de Gran Canaria\\
{\tt\small yopedro1989@gmail.com}
}

\maketitle
\thispagestyle{empty}

\begin{abstract}
  Color based re-identification methods usually rely on a distance function to measure the similarity between individuals.  In this paper we study the behavior of several histogram distance measures in different color spaces. We wonder whether there is a particular histogram distance measure better than others, likewise also, if there is a color space that present better discrimination features. Several experiments are designed and evaluated in several images to obtain  measures against various color spaces. We test in several image databases. A measure ranking is generated  to calculate the area under the CMC, this area is the indicator used to evaluate which distance measure and color space present the best performance for the considered databases. Also, other parameters such as the image division in horizontal stripes and number of histogram bins, have been studied.

\end{abstract}

\section{Introduction}

Nowadays thanks to cheaper sensors and processors for video cameras, surveillance camera networks are widely present. These cameras can be useful in locating disappeared persons, tracking thieves, incident detection, traffic control, etc. The scenarios can be very varied, but may be roughly  divided into indoor and outdoor areas such as a hospital or highway respectively. These monitoring systems accumulate a huge amount of information that must be processed. Human processing of the visual information gathered by the camera network may be unfeasible in certain scenes, so intelligent systems arise for re-identification \cite{recogletters}.

According to \cite {reident}, "the concept of re-identification is defined as the fundamental task for a system of distributed cameras or not, with an association of people is through the images captured by these at a certain location and time". In a re-identification problem we have the probe that is the individual that must be matched with the previously seen individuals, i.e. the gallery. As shown in equation (\ref{ecuaREID}), the task of re-identificacion is formally defined, as follows:

\begin{equation}
 T = arg min_{T_i} D(T_i,Q) , T_i \in \tau
 \label{ecuaREID}
 \end{equation}
where $D()$ is a similarity measure, that matches the probe $Q$ with the n candidates from the gallery $ \tau = \left\lbrace   \right\rbrace $.

There are different parameters that must be considered in the re-identification problem. To describe individuals, we must take into account the descriptors, that are calculated for each individual. These descriptors can be based on color, shape, texture, position or biometric features. 

Two different set of characteristics are commonly extracted from the images for this problem, part-based body models and features \cite{DBLP:journals/corr/Satta13}. Part-based body models represents how the body  is divided to obtain the features in this areas. There are different models to extract information as may be Fixed Part Models, Adaptative Part Models and Learned Part Models.  Fixed Part Models divide the individual in few horizontal and fixed stripes that represent the head, torso and legs \cite{avraham2012learning, 5692588}. Other models use euristic methods and classifiers to identify parts of the body \cite{canny, cris, ghei}. On the other hand,  the image is described by one or more descriptors that represent the features of the image, that could be global or local features. Global features are the characteristics that represent the image or a part of it, they could be color histograms, textures or edges \cite{citeulike:1737781, Schmid01cvpr}. Local features are related to pixels in a small area of the image, the most extended methods are SIFT (Scale Invariant Feature Transform), SURF (Speeded-Up Robust Features) or LBP (Local Binary Pattern) \cite{Lowe:2004:DIF:993451.996342, Bay:2008:SRF:1370312.1370556, Ojala:2002:MGR:628329.628808}.

 We focus on the feature set, specifically in descriptors based on color.  There are multiple color spaces with their respective characteristics in relation to the luminance and chrominance.

Color spaces that are often used in the field of computer vision are:

\begin{itemize}
\item RGB (Red Green Blue) is the typical color space found in many devices, it conform 3 channels (red, green and blue) where the luminance and chrominance are not found separately. It is not a perceptually uniform space. Distant colors are not perceived as such, and vice versa.
\item HSV (Hue Saturation Value) is a color space that consists of three channels that characterize the hue (H), saturation (S) and value (V). It is a non linear transformation from RGB.

    

\item CIELAB  was defined by CIE (Commission Internationale de l'éclairage), L (Lightness) and A and B for the color-opponent dimensions. It is a color space that aims to be a linear color space. 

%
\item YCbCr is a color space that is made up  by a luminance component (Y) and two color components (Cb and Cr), that represent the chrominance in blue and red. 
%
\end{itemize}

In some problems, the image histogram provides enough information to describe an image. This method aims to obtain only color information of the image lacking of spatial information. A histogram represents the number of occurrences of the pixel values in the image, it is defined formally in equation (\ref {histecua}). A histogram is made up of a certain number of bins. Each bin would be the intervals in which is divided the whole range of measurement values representing the histogram.
\begin{equation}
H_i(A) = \sum_{j=1}^n c_{ij} \quad \textbf{where } c_{ij}  =        \left\{
  \begin{array}{ll}
   1 & \textbf{If } a_j = x_i\\
   0 & \textbf{otherwise}
  \end{array}
\right.
\label{histecua}
\end{equation} 

When working with histograms, a good practice is to perform a normalization, that is usually adopted following an approach to obtain a description similar to a probability function, see equation (\ref {funcPROB}). This preprocessing task is performed to bring in a common plane any distribution regardless of the image size. There is a need to compare histograms to know how similar they are, as identified in equation (\ref{ecuaREID}) \cite{4}, for this reason, the litareture has defined different types of distance measures for histograms.

\begin{equation}
H'(A) = \frac{H(A)}{\sum_{i=1}^{n}A_i} \quad \textbf{ where }\sum_{i=1}^{n}H(A) = 1
\label{funcPROB}
\end{equation}

A distance $ D (x, y) $ is defined in a space of dimension $ \mathbb{R}^n $, $ d: \mathbb{R}^n \times \mathbb {R}^n \rightarrow \mathbb{R} $, it must satisfy the following properties:

\begin{itemize}
  \item $D(x,y) \geq 0$ 
   \item $D(x,x) = 0$
  \item $D(x,y) = D(y,x)$
  \end{itemize}
  If it also satisfy:
\begin{itemize}
  \item $D(x,y)=0 \textbf{ iff } x = y$
  \item $D(x,y) \leq D(x,k) + D(k,y)$
\end{itemize}
It is said to be a metric distance.

There are two groups of distance measures for histograms: bin to bin and cross-bin. The first group focuses on the comparison of the content of the bin with the corresponding bin of the second histogram, it does not exploit the information of neighborhood bins. The second group emphasizes the neighborhood values of the bin to be treated. Some  distance measures that are commonly used between histograms are the following:
\begin{itemize}
\item Bin to bin measures:
\begin{itemize}
 \item Bhattacharyya \cite{bin1} measures the similarity of two probability distributions. It has a computational complexity O$(n)$. 
  \begin{equation}
  D(x,y)=1-\sqrt{\sum_{i=1}^n\frac{\sqrt{x_iy_i}}{\sqrt{\sum_{i=1}^nx_i\sum_{i=1}^ny_i}}}
  \label{equa:bat}
\end{equation}   
\item Chi Square \cite{doi:10.1117/12.2042359} has statistical origin. It has a computational complexity O$(n)$.
   \begin{equation}
     D(x,y)=\frac{1}{2}\sum_{i=1}^n\frac{(x_i-y_i)^2}{(x_i+y_i)} 
     \label{equa:chi}
  \end{equation}
  \item Correlation \cite{histcom} is a measure that refers to a statistical relationship that involve dependence. It has a computational complexity O$(n)$.
  \begin{equation}
  	D(x,y)=\frac{\sum_{i=1}^n(x_i-\bar{x})(y_i-\bar{y})}{\sqrt{\sum_{i=1}^n(x_i - \bar{x})^2 \sum_{i=1}^n(y_i-\bar{y})^2}}
  	\label{equa:correl}
 \end{equation}   
 \item Intersection \cite{doi:10.1117/12.2042359} is a measure that comes from the intersection of the two histograms. This measure has high performance. It has a computational complexity O$(n)$.
  \begin{equation}
  D(x,y)=\sum_{i=1}^n min(x_i,y_i)
  \label{equa:inter}
  \end{equation}
\item KL (Kullback-Leibler) \cite{1211407} divergence is a measure that has the origin in the area of information theory. This measure  does not verify the symmetry property. In the second histogram bins can not have zero value as this causes an uncertainty. It has a computational complexity O$(n)$. 
  \begin{equation}
  D(x,y)=\sum_{i=1}^nx_i\log\frac{x_i}{y_i}
  \label{equa:kl}
  \end{equation}
\end{itemize}
\item Cross-bin distances:
	\begin{itemize}
	\item Earth mover's distance (EMD) \cite{doi:10.1117/12.2042359} is a measure which is defined by the minimum cost that we should pay to transform a histogram into another. It has a computational complexity O($n^ 3 \log n$).
	\begin{equation}
  \begin{split}
  D(x,y) = min_{F=\{F_{ij}\}}\frac{\sum_{i,j}F_{ij}D_{ij}}{\sum_iF_{ij}}\\
  s.t: \quad \sum_j F_{ij}\leq x_i \quad \sum_i F_{ij} \leq y_j\\
  \quad\quad \sum_{i,j}F_{ij} = min(\sum_ix_i,\sum_jy_j)\\
  F_{ij}\geq 0
  \end{split}
  \label{equa:emd}
  \end{equation}
  \item Mahalanobis \cite{5647609} is a measure which is the distance between a point and a distribution. It has a computational complexity O$\left( n^2 \right)$.
   \begin{equation}
	\begin{split}
  D(x,y) = \sqrt{(x-y)S^{-1}(x-y)} \\
  \end{split}
 \label{equa:maha} 
  \end{equation}   
  where  $S$ is the covariance matrix.
	\end{itemize}
\end{itemize}
\section{Methodology}
\subsection{Databases}
\begin{table}[h!]
  \centering
    \begin{tabular}{|m{2.1cm}|m{1.7cm}|m{1.7cm}|m{1cm}|}
    \hline
        \centering Database &\centering Number of individuals &\centering Resolution & Image \bigstrut\\
    \hline
    \centering CAVIAR4REID & \centering 1220  &\centering From 17x39 to 72x141 &    \includegraphics[width=0.05\textwidth]{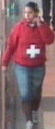}\bigstrut[t]\\
   \centering i-LIDS &\centering 477   &\centering From 32x76 to 115x294 &    \includegraphics[width=0.05\textwidth]{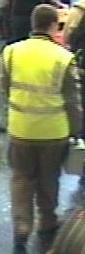}\\
    \centering VIPeR &\centering 1264  & \centering 48x128 &  \includegraphics[width=0.05\textwidth]{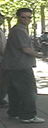}  \\
    \centering QMUL GRID &\centering 1275  &\centering 48x128 &  \includegraphics[width=0.05\textwidth]{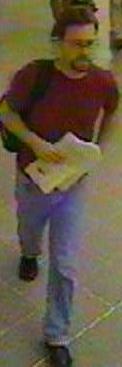}  \bigstrut[b]\\
    \hline
    \end{tabular}%
  \label{tab:compBD}%
   \caption{Comparative of databases that are used with number of individuals, resolution (pixels) and sample image.}
\end{table}
To perform the experiment, we consider several databases. Each database has its own characteristics, illumination conditions, capture sensor, noise present, etc. We use CAVIAR4REID \cite{caviar}, i-LIDS \cite{ilids}, VIPeR \cite{viper} and QMUL GRID \cite{grid}. Some of their charectaristics are shown in table \ref{tab:compBD}.

\subsection{Division by stripes}

\begin{figure}[h!]
 \centering
  \includegraphics[width=0.37\textwidth]{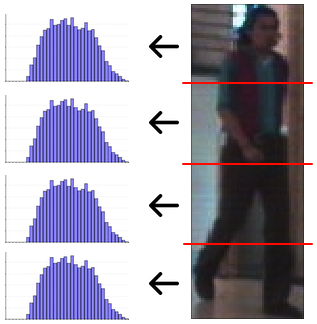}
  \caption{Individual image sample stripe division.}
  \label{est}
 \end{figure}

The division of the individual image in stripes \cite{avraham2012learning} or sections, see figure \ref{est}, can be a methodology to obtain good results. The results are improved because a distribution of local color is obtained, not being affected by noise. For instance, we could obtain the information of the head, torso and legs. Furthermore, smaller areas could be weighted with lower values and vice versa. However, the number of histogram bins must be carefully configured because some colors could not be present in the  histogram and that could produce indeterminations in the distance computation.

\subsection{Experiment design}
It is interesting to know if there is any histogram distance measure that outperforms the others, to study the behavior of the measures we will consider different parameters that can affect them. These features are:

\begin{itemize}
 \item Distance measures: Bhattacharyya, Chi Square, Correlation, EMD, Intersection, Mahalanobis and Kullback-Leibler
  \item Color spaces: RGB, HSV and CIELAB
  \item Number of stripes of the image division: complete image, 5 stripes, 10 stripes and 25 stripes
  \item Number of bins for the histogram: 16, 32, 64 and 128
  \item Databases: CAVIAR4REID, i-LIDS, VIPeR and QMUL GRID
  \end{itemize}  
  
  \subsection{Evaluation}
  The area under CMC (Cumulative Match Curve) is used as performance indicator. CMC is widely used  in re-identification as performance measure   \cite{1544394}. To calculate the CMC, we must rank in decreasing order of similarity the individuals of the gallery for each probe. Later, the position of each respective probe is accumulated. Finally each element is divided by the number of probes and the graphics are generated from theses values as the accumulation of the above elements. 
  
  \section{Results}
  
 In order to eliminate possible indeterminations during the distances computation. We have firstly modified Chi Square and KL (Kullback-Liebler) distances. For the first distance, it can be observed in the equation (\ref{equa:chi}) that if the denominator is zero the result would be infinite, this situation can occur when two bins are being processed with null values. Our approach  to eliminate this uncertainty has been to discard these bins in the calculation. For the second measure, looking at the corresponding KL distance  equation (\ref{equa:kl}), there can not be bins with zero because two situations may occur: the denominator is zero and we would get as infinite value or the numerator is zero and as a result would obtain $ \ ln (0) $ that is equal to infinity. The solution is to discard from the analysis the pairs of bins that when one of them compared is equal to null.
 
Depending on the distance, the higher similarity can correspond to the lower value or the higher value of them. Therefore, the equations  (\ref{equa:correl}, \ref{equa:inter}, \ref{equa:kl})  that correspond  to Correlation, Intersection and KL distances respectively have been modified in other to get the lower the value, the higher the similarity. This comes from the need to share a similar representation range to all the distance measures. To solve these problems, we  have changed the conditions of the equations  Correlation and Intersection as shown in equation (\ref{eq:modRango}), in addition, to solve the KL distance has been added the following conditions, Equation (\ref{eq:modRango2}).
 
 If distance $=$ \{Correlation or Intersection\}
\begin{equation}
distance_f(x, y) = - distance(x, y)
\label{eq:modRango}
\end{equation}

If distance $=$ KL
\begin{equation}
\left\lbrace 
\begin{matrix}
 KL_f(x,y)=KL(x,y) & \textbf{ If } KL(x,y) \geq 0 \\
  KL_f(x,y)=-KL(x,y)& \textbf{ If } KL(x,y) < 0 
\end{matrix}
\right.
\label{eq:modRango2}
\end{equation}

\begin{figure}[h!]
 \centering
  \includegraphics[width=0.45\textwidth]{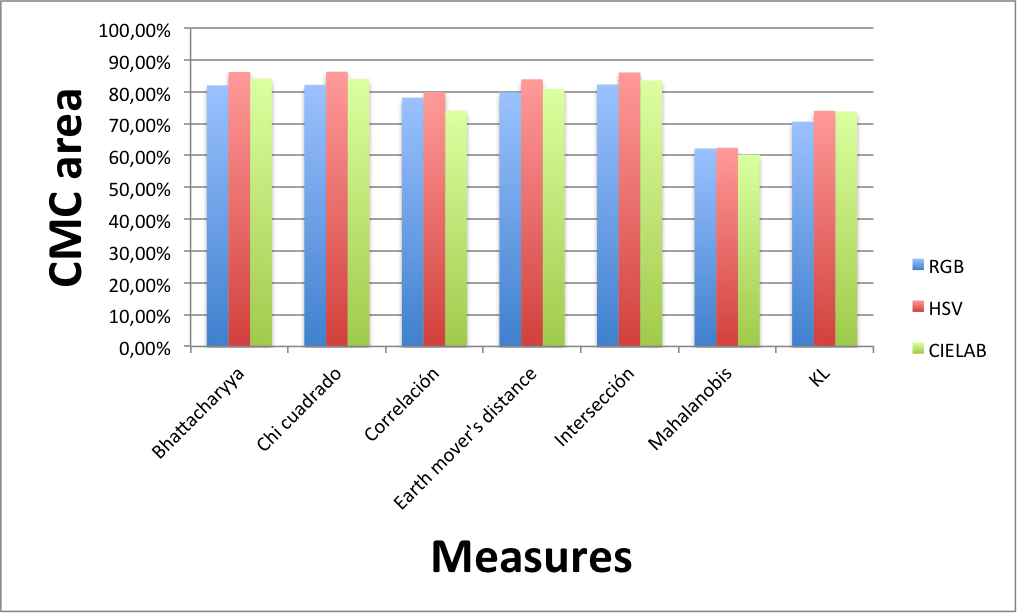}
  \caption{CMC area for distances and color space.}
  \label{graph:distColor}
 \end{figure}

As the graphical representation of the CMC is difficult to use to compare results, we have obtained the CMC area generated by the average number of bins, number of stripes and databases. In this way we visualize the data as shown in Figure \ref{graph:distColor}, where the abscissa groups the results of CMC areas for each distance and color spaces. Those distances that provide better results are Bhattacharyya, Chi Square and Intersection. These three measures are bin to bin distances, and agree with the same between color spaces. The color space with best performance is the HSV, circumstance that may be related be related with  the separation in components, chrominance and luminance.

\begin{figure}[h!]
 \centering
  \includegraphics[width=0.45\textwidth]{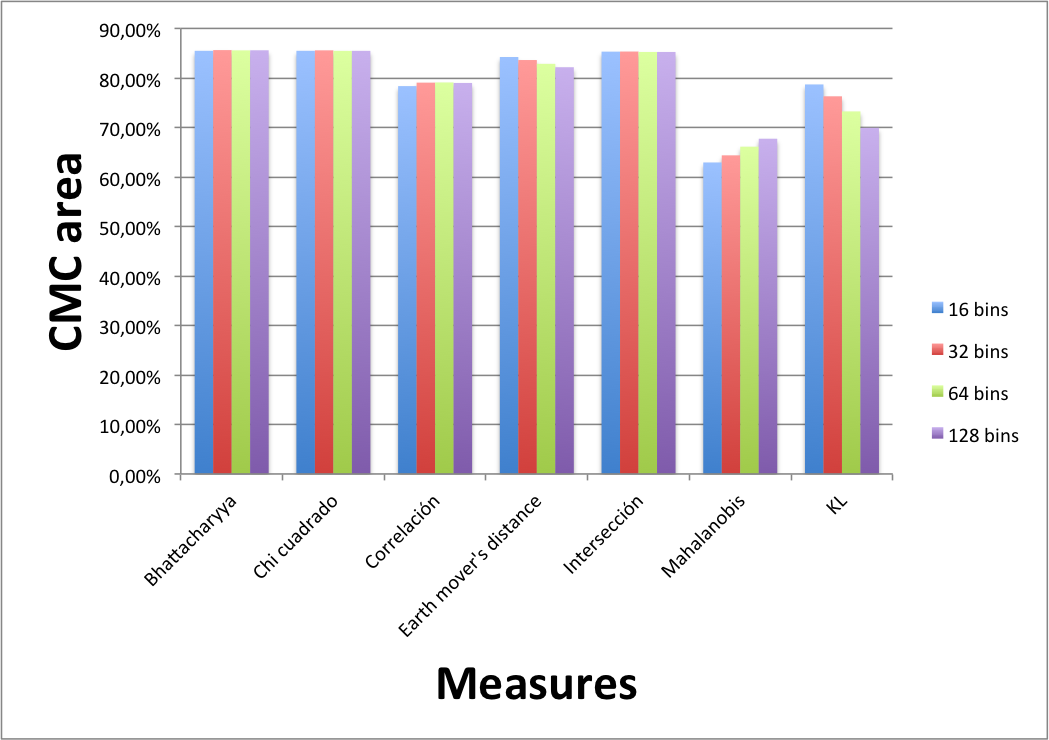}
  \caption{CMC area for distances and number of bins.}
  \label{graph:distBin}
 \end{figure}

To observe the effect of the different number of bins, we obtain the CMC area generated by the average of color space, number of stripes and space databases. Again we visualize the results in Figure \ref {graph:distBin}, where the abscissa groups the results of CMC areas for each distance with different number of bins. Distances that obtain good results are Bhattacharyya, Chi Square and Intersection. It could be emphasized that the reduction of the histogram using a big bin size, provides similar results using smaller bins. It would be interesting to  use larger bins because calculations may be computed faster, as the histogram has less number of bins. 
We could discuss the behavior that is perceived in the KL measure. KL measure gets worst results with higher number of bins, this is due to the approach we have used to resolve the uncertainties of KL. Increasing the number of bins will increase the probability that a color does not appear in a stripe and will form holes in the histogram, these voids will be zero, which implies that they will not be processed when the calculation of the distance with the other color histogram, although other color histogram have values other than zero. So, reducing the number of bins, we lose information to be processed, a possible solution is to use the Jeffrey divergence   \cite{Rubner:2000:EMD:365875.365881}. In conclusion, the use of a small number of bins provides better results, but does not obtain a significant improvement. We propose an initial configuration to a problem to use 16 or 32 bins for the histogram, not to lose too much information.
 
\begin{figure}[h!]
 \centering
  \includegraphics[width=0.45\textwidth]{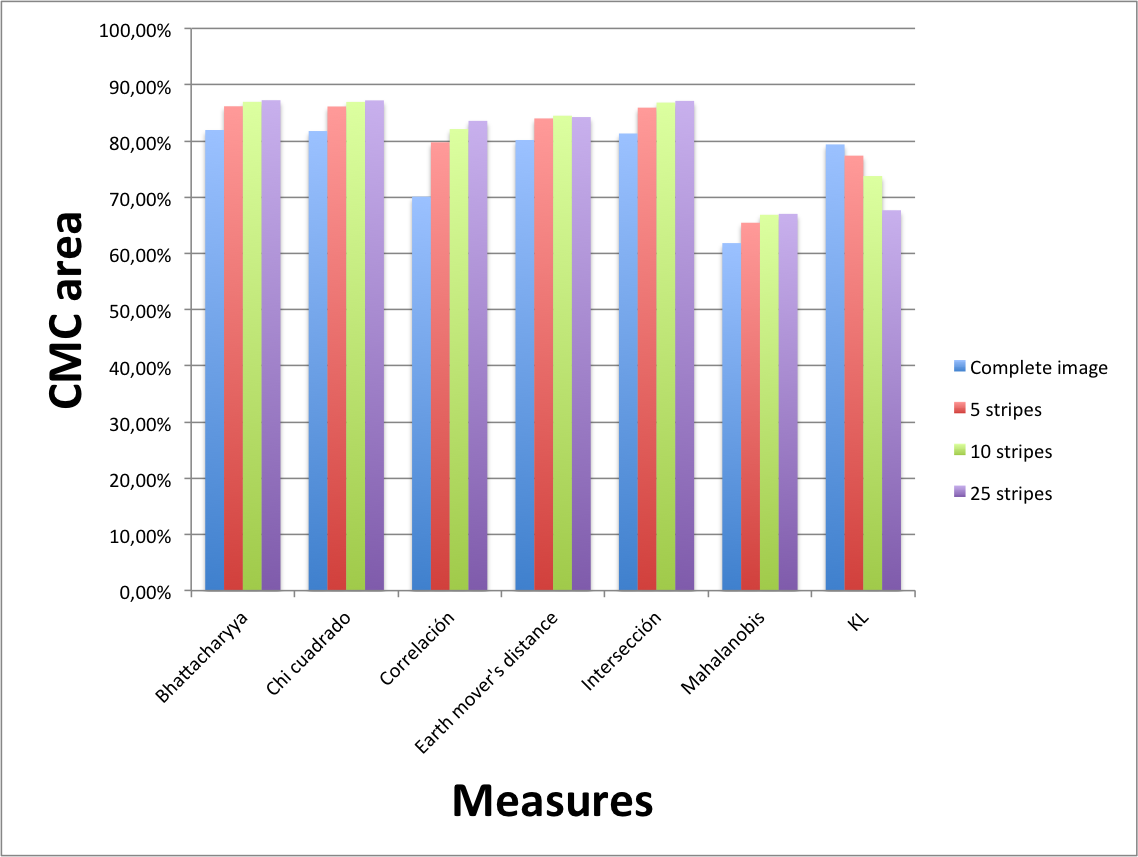}
    \caption{CMC area for distances and number of stripes.}
  \label{graph:distFranjas}
 \end{figure}

Observing the influence related to the number of stripes, we obtain the CMC area generated by the average of color space, number of bins and image  databases. Again we visualize the results in Figure \ref {graph:distFranjas}, where the abscissa groups the results of CMC areas for each distance with different configurations of stripes. 
 Distances that provide better results are Bhattacharyya, Chi Square and Intersection. Dividing the image into stripes significantly improves the results, but comes to a point at which stripes excess makes the results worse. This is because the higher the number of stripes, the less pixels they have. Having a lot of stripes we will have many histograms with null  bins values. On the other hand, making use of one histogram for the image, detailed information of the color distribution in certain areas of the image is lost.  KL distance behaves in an abnormal manner compared with the rest of distances. This is because as described before, important information has been discarded during distance computation. We propose to use as the initial configuration 5 or 10 stripes, as both configuration exhibit best results and between them do not differ significantly.
 
\begin{figure}[h!]
 \centering
  \includegraphics[width=0.45\textwidth]{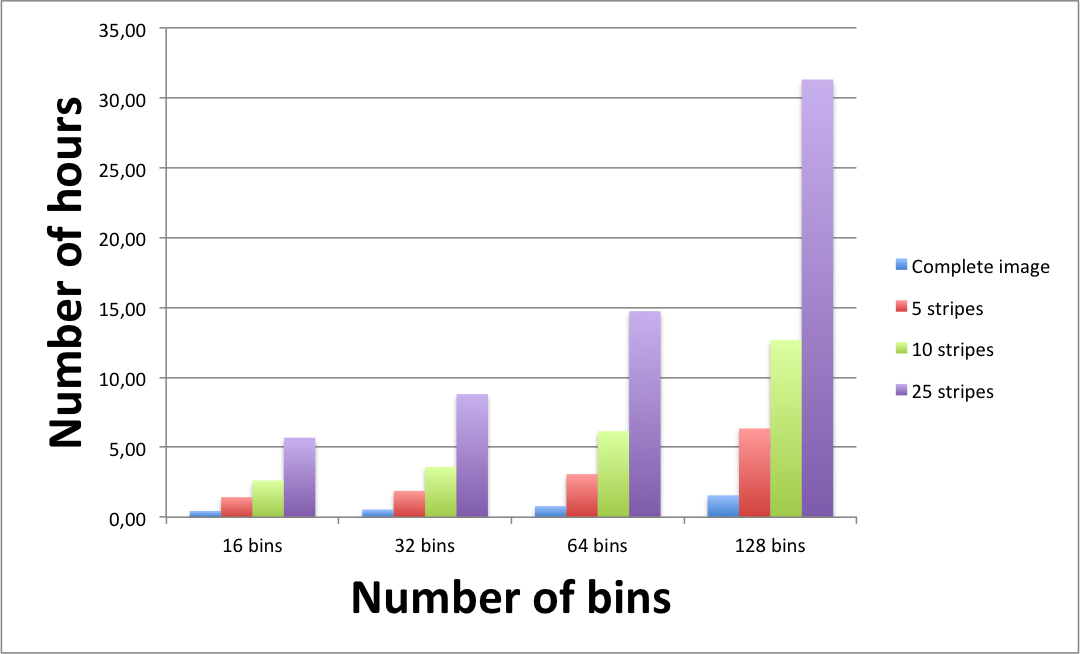}
    \caption{Performance in hours for number of bins and number of stripes.}
  \label{graph:rendimiento}
 \end{figure}

It is interesting to check the processing time, we should have a commitment in relation to time and results at the time of re-identify. Figure \ref{graph:rendimiento} shows testing execution time measued in hours, where we refer to the number of bins and the number of stripes. These results refer to the total time to execute the 4 databases with 3 color spaces, 7 distance measures and 16 settings for the number of bins and the number of stripes. 
Increasing the number of bins and the number of stripes affects clearly the runtime of the tests. It can be seen that there is a relationship with the number of stripes, where each bin increases approximately linearly. Unlike, the number of bins does not appreciate a clear relationship.
 
 \section{Conclusion}

We have chosen to study the problem of re-identification based on appearance, using color histograms for image representation. We have evaluated different distance measures for comparing histograms for various  color spaces with different number of bins configurations and number of stripes in the image.
 
 We propose an initial configuration to solve the problem of re-identification, Table \ref{tab:addlabel}, this configuration does not assure to be the best configuration for a specific problem, but this configuration has a high probability of success. After performing  and analysing the experiments, we have obtained the following conclusions.
 
 We propose Bhattacharyya, Chi Square and Intersection as a first approach to solving the problem, because these measures achieve good performance  and very similar results for all the settings. HSV is the color space that reported better results by far. It can be due to the separation into luminance and chrominance. It is quite useful when we are dealing scenes with changing illumination conditions.
 
To configure the number of bins in the histogram, we proposed to use as a starting value between 16 and 32 bins. This is because a large number of bins in the histogram generates noise because the image does not contain the full range of colors. Besides adding a computational cost when data are processed. Finally, we have chosen to divide the image between 5 and 10 stripes that are the configuration which performed best because complete image does not bring knowledge on specific areas. Instead, making excessive divisions generate noise in the histograms.
 \begin{table}[htbp]
  \centering
    \begin{tabular}{|m{2.5cm}|>{\centering\arraybackslash}m{4.5cm}|}
    \hline
          & Preference configuration \\
    \hline
    Distance & Bhattacharyya, Chi Square and Intersection\\
    Color Space & HSV \\
    Number of bins & 16 and 32 \\
    Number of stripes & 5 and 10 \\
    \hline
    \end{tabular}%
    \caption{Proposed initial configuration.}
  \label{tab:addlabel}%
\end{table}

As future work, we plan to use more color spaces for perform the tests. It would also be convenient to test a larger number of databases. Paper \cite {survDB}  exposes a database which is the agglomeration of multiple sets of images, where  images with multiple characteristics are included. Furthermore, for the study would be favorable to use a larger number of distance measures, as may be the divergence of Jeffrey to resolve the KL uncertainties. Finally, we could make a new experiment based on our proposed initial configuration and using different image sizes. This is due to the high computational cost of repeating all experiments from the beginning.

{\small
\bibliographystyle{ieee}
\bibliography{referencias}
}

\end{document}